\pdfoutput=1
\documentclass[11pt]{article}
\usepackage{naacl2021}

\usepackage{times}
\usepackage{latexsym}

\usepackage[T1]{fontenc}

\usepackage[utf8]{inputenc}
\usepackage{microtype}
\usepackage{graphicx}
\usepackage{booktabs}
\usepackage{adjustbox}
\usepackage{color,soul,colortbl}
\usepackage{caption}
\usepackage[hang,flushmargin]{footmisc}

\usepackage{amssymb}
\usepackage{amsbsy}

\AtBeginDocument{}

\definecolor{hexpurple}{HTML}{d9d2e8}
\definecolor{hexgreen}{HTML}{d9ead4}

\title{SIGTYP 2021 Shared Task: Robust Spoken Language Identification}

\newcommand{\jhu}{\diamond}
\newcommand{\saar}{\ddag}
\newcommand{\hse}{\triangleleft}
\newcommand{\mila}{\Box}
\newcommand{\bra}{\bullet}
\newcommand{\ethz}{\odot}
\newcommand{\unimelb}{\sharp}
\newcommand{\iling}{\triangleright}

\author{
Elizabeth Salesky$^{\jhu}$\thanks{~~Equal contribution}~~~
Badr M. Abdullah$^{\saar}$\footnotemark[1]~~~
Sabrina J. Mielke$^{\jhu}$\footnotemark[1]
\\
\textbf{
Elena Klyachko$^{\hse, \iling}$~~~
Oleg Serikov$^{\hse}$~~~
Edoardo Ponti$^{\mila}$
}
\\
\textbf{
Ritesh Kumar$^{\bra}$~~~
Ryan Cotterell$^{\ethz}$~~~
Ekaterina Vylomova$^{\unimelb}$
}
\\
  $^{\jhu}$Johns Hopkins University~~~
  $^{\saar}$Saarland University~~~
  $^{\hse}$Higher School of Economics
\\
  $^{\iling}$The Institute of Linguistics RAS
  $^{\mila}$Mila/McGill University Montreal~~~
\\
  $^{\bra}$Bhim Rao Ambedkar University
  $^{\ethz}$ETH Zürich~~~
  $^{\unimelb}$University of Melbourne~~~
}

\begin{document}
\maketitle
\begin{abstract}
While language identification is a fundamental speech and language processing task, for many languages and language families it remains a challenging task. 
For many low-resource and endangered languages this is in part due to resource availability: where larger datasets exist, they may be single-speaker or have different domains than desired application scenarios, demanding a need for domain and speaker-invariant language identification systems. 
This year’s shared task on robust spoken language identification sought to investigate just this scenario: systems were to be trained on largely single-speaker speech from one domain, but evaluated on data in other domains recorded from speakers under different recording circumstances, mimicking realistic low-resource scenarios. 
We see that domain and speaker mismatch proves very challenging for current methods which can perform above 95\% accuracy in-domain, which domain adaptation can address to some degree, but that these conditions merit further investigation to make spoken language identification accessible in many scenarios. 
\end{abstract}

\section{Introduction}
Depending on how we count, there are roughly 7000 languages spoken around the world today. The field of linguistic typology is concerned with the study and categorization of the world's languages based on their linguistic structural properties \cite{comrie1988linguistic, crofttypology}. While two languages may share structural properties across some typological dimensions, they may vary across others. For example, two languages could have identical speech sounds in their phonetic inventory, yet be perceived as dissimilar because each has its own unique set of phonological rules governing possible sound combinations. This leads to tremendous variation and diversity in speech patterns across the world languages \cite{tucker2020introduction}, the effects of which are understudied across many downstream applications due in part to lack of available resources. Building robust speech technologies which are applicable to any language is crucial to equal access as well as the preservation, documentation, and categorization of the world's languages, especially for endangered languages with a declining speaker community. 

Unfortunately, robust (spoken) language technologies are only available for a small number of languages, mainly for speaker communities with strong economic power. The main hurdle for the development of speech technologies for under-represented languages is the lack of high-quality transcribed speech resources (see \citet{joshi2020state} for a detailed discussion on linguistic diversity in language technology research). The largest multilingual speech resource in terms of language coverage is the CMU Wilderness dataset \cite{black2019wilderness}, which consists of read speech segments from the Bible in $\sim$700 languages. Although this wide-coverage resource provides an opportunity to study many endangered and under-represented languages, it has a narrow domain and lacks speaker diversity as the vast majority of segments are recorded by low-pitch male speakers. It remains unclear whether such resources can be exploited to build generalizable speech technologies for under-resourced languages. 

Spoken language identification (SLID) is an enabling technology for multilingual speech communication with a wide range of applications.  Earlier SLID systems addressed the problem using the phonotactic approach whereby generative models are trained on sequences of phones transduced from the speech signal using an acoustic model \cite{lamel1994language, li2005phonotactic}. Most current state-of-the-art SLID systems are based on deep neural networks which are trained end-to-end from a spectral representation of the acoustic signal (e.g., MFCC feature vectors) without any intermediate symbolic representations \cite{lopez2014automatic, gonzalez2014automatic}. In addition to their ability to effectively learn to discriminate between closely related language varieties \cite{gelly2016language, shon2018convolutional}, it has been shown that neural networks can capture the degree of relatedness and similarity between languages in their emergent representations \cite{abdullah-etal-2020-rediscovering}. 

Several SLID evaluation campaigns have been organized in the past, including the NIST Language Recognition Evaluation \cite{lee20162015, sadjadi20182017}, focusing on different aspects of this task including closely related languages, and typically used conversational telephone speech. However, the languages were not sampled according to typologically-aware criteria but rather were geographic or resource-driven choices. Therefore, while the NIST task languages may represent a diverse subset of the world's languages, there are many languages and language families which have not been observed in past tasks. In this shared task, we aim to address this limitation by broadening the language coverage to a set of typologically diverse languages across seven languages families. We also aim to assess the degree to which single-speaker speech resources from a narrow domain can be utilized to build robust speech language technologies. 

\section{Task Description}

While language identification is a fundamental speech and language processing task, it remains a challenging task, especially when going beyond the small set of languages past evaluation has focused on. Further, for many low-resource and endangered languages, only single-speaker recordings may be available, demanding a need for domain and speaker-invariant language identification systems.

We selected 16 typologically diverse languages, some of which share phonological features, and others where these have been lost or gained due to language contact, to perform what we call robust language identification: systems were to be trained on largely single-speaker speech from one domain, but evaluated on data in other domains recorded from speakers under different recording circumstances, mimicking more realistic low-resource scenarios. 

\begin{table*}[th]
\centering
\resizebox{\textwidth}{!}{%
\begin{tabular}{llllllll}
\toprule
\textbf{ISO} & \textbf{Wilderness ID} & \textbf{Language name} & \textbf{Family} & \textbf{Genus}          & \textbf{Macroarea} & \textbf{Train} & \textbf{Eval} \\
\midrule
kab & KABCEB          & Kabyle        & Afro-Asiatic  & Berber                  & Africa    & Wilderness & CV    \\
iba & IBATIV          & Iban          & Austronesian  & Malayo-Sumbawan         & Papunesia & Wilderness & SLR24 \\
ind & INZTSI          & Indonesian    & Austronesian  & Malayo-Sumbawan         & Papunesia & Wilderness & CV    \\
sun & SUNIBS          & Sundanese     & Austronesian  & Malayo-Sumbawan         & Papunesia & Wilderness & SLR36 \\
jav & JAVNRF          & Javanese      & Austronesian  & Javanese                & Papunesia & Wilderness & SLR35 \\
eus & EUSEAB          & Euskara       & Basque        & Basque                  & Eurasia   & Wilderness & CV    \\
tam & TCVWTC          & Tamil         & Dravidian     & Southern Dravidian      & Eurasia   & Wilderness & SLR65 \\
kan & ERVWTC          & Kannada       & Dravidian     & Southern Dravidian      & Eurasia   & Wilderness & SLR79 \\
tel & TCWWTC          & Telugu        & Dravidian     & South-Central Dravidian & Eurasia   & Wilderness & SLR66 \\
hin & HNDSKV          & Hindi         & Indo-European & Indic                   & Eurasia   & Wilderness & SS    \\
por & PORARA          & Portuguese    & Indo-European & Romance                 & Eurasia   & Wilderness & CV    \\
rus & RUSS76          & Russian       & Indo-European & Slavic                  & Eurasia   & Wilderness & CV    \\
eng & EN1NIV          & English       & Indo-European & Germanic                & Eurasia   & Wilderness & CV    \\
mar & MARWTC          & Marathi       & Indo-European & Indic                   & Eurasia   & Wilderness & SLR64 \\
cnh & CNHBSM          & Chin, Hakha   & Niger-Congo   & Gur                     & Africa    & Wilderness & CV   \\   
tha & THATSV          & Thai          & Tai-Kadai     & Kam-Tai                 & Eurasia   & Wilderness & CV    \\
\bottomrule
\end{tabular}
}%
\caption{Provided data with language family and macroarea information. \textbf{ISO} shows ISO 639-3 codes. Training data (\textbf{Train}) for all languages is taken from CMU Wilderness dataset; validation and evaluation data (\textbf{Eval}) is derived from multiple data sources. }
\label{tab:provided-data}
\end{table*}

\subsection{Provided Data}

To train models, we provided participants with speech data from the CMU Wilderness dataset \cite{black2019wilderness}, which contains utterance-aligned read speech from the Bible in 699 languages,\footnote{Data source: \url{bible.is}} but predominantly recorded from a single speaker per language, typically male. 
Evaluation was conducted on data from other sources---in particular, multi-speaker datasets recorded in a variety of conditions, testing systems' capacity to generalize to new domains, new speakers, and new recording settings. 
Languages were chosen from the CMU Wilderness dataset given availability of additional data in a different setting, and include several language families as well as more closely-related challenge pairs such as Javanese and Sundanese. 
These included data from the Common Voice project \cite[CV;][]{ardila-etal-2020-common-voice} which is read speech typically recorded using built-in laptop microphones; 
radio news data \cite[SLR24;][]{SLR24-1,SLR24-2}; 
crowd-sourced recordings using portable electronics \cite[SLR35, SLR36;][]{SLR35-36}; 
cleanly recorded microphone data \cite[SLR64, SLR65, SLR66, SLR79;][]{SLR64-65-66-79};
and a collection of recordings from varied sources \cite[SS;][]{SS}. 
\autoref{tab:provided-data} shows the task languages and their data sources for evaluation splits for the robust language identification task. 

We strove to provide balanced data to ensure signal comes from salient information about the language rather than spurious correlations about e.g. utterance length. 
We selected and/or trimmed utterances from the CMU Wilderness dataset to between 3 to 7 seconds in length. 
Training data for all languages comprised 4,000 samples each. 
We selected evaluation sources for validation and blind test sets to ensure no possible overlap with CMU Wilderness speakers. 
We held out speakers between validation and test splits, and balanced speaker gender within splits to the degree possible where annotations were available. 
We note that the Marathi dataset is female-only. 
Validation and blind test sets each comprised 500 samples per language.  
We release the data as derivative MFCC features. 

\section{Evaluation}

The robust language identification shared task allowed two kinds of submissions: first, \textit{constrained} submissions, for which only the provided training data was used; and second, \textit{unconstrained} submissions, in which the training data may be extended with any external source of information (e.g. pre-trained models, additional data, etc.).

\subsection{Evaluation Metrics}

We evaluate task performance using precision, recall, and F$_1$. 
For each metric we report both micro-averages, meaning that the metric average is computed equally-weighted across all samples for all languages, and macro-averages, meaning that we first computed the metric for each language and then averaged these aggregates to see whether submissions behave differently on different languages. 
Participant submissions were ranked according to macro-averaged F$_1$.

\subsection{Baseline}

For our baseline SLID system, we use a deep convolutional neural network (CNN) as sequence classification model. The model can be viewed as two components trained end-to-end: a segment-level feature extractor ($f$) and a language classifier ($g$). Given as input a speech segment parametrized as sequence of MFCC frames $\mathbf{x}_{1:T} = (\mathbf{x}_1, \dots, \mathbf{x}_T) \in \mathbb{R}^{k \times T}$, where $T$ is the number of frames and $k$ is the number of the spectral coefficients, the segment-level feature extractor first transforms $\mathbf{x}_{1:T}$ into a segment-level representation as $\mathbf{u} = f(\mathbf{x}_{1:T}; \boldsymbol{\theta}_f) \in \mathbb{R}^d$. Then, the language classifier transforms $\mathbf{u}$ into a logit vector $\mathbf{\hat{y} \in \mathbb{R}^{|\mathcal{Y}|}}$, where $\mathcal{Y}$ is the set of languages,  through a series of non-linear transformations as $\mathbf{\hat{y}} = g(\mathbf{u}; \boldsymbol{\theta}_g)$. The logit vector $\mathbf{\hat{y}}$ is then fed to a softmax function to get a probability distribution over the languages.

The segment-level feature extractor consists of three 1-dimensional, temporal convolution layers with 64, 128, 256 filters of widths 16, 32, 48 for each layer and a fixed stride of 1 step. Following each convolutional operation, we apply batch normalization, ReLU non-linearity, and unit dropout with probability which was tuned over $\{0.0, 0.4, 0.6\}$. We apply average pooling to downsample the representation only at the end of the convolution block, which yields a segment representation $\mathbf{u} \in \mathbb{R}^{256}$. The language classifier consists of 3 fully-connected layers (256 $\rightarrow$ 256 $\rightarrow$ 256 $\rightarrow$ 16), with a unit dropout with probability $0.4$ between the layers, before the softmax layer. The model is trained with the ADAM optimizer with a batch size of 256 for 50 epochs. We report the results of the best epoch on the validation set as our baseline results.

\subsection{Submissions}

We received three constrained submissions from three teams, as described below.

\textbf{Anlirika} \cite[composite]{anlirika2021sigtyp}
The submitted system (constrained) consists of several recurrent, convolutional, and dense layers.
 The neural architecture starts with a dense layer that is designed to remove sound harmonics from a raw spectral pattern. This is followed by a 1D convolutional layer that extracts audio frequency patterns (features). Then the features are fed into a stack of LSTMs that focuses on `local' temporal constructs. The output of the stack of LSTMs is then additionally concatenated with the CNN features and is fed into one more LSTM module. Using the resulting representation, the final (dense) layer evaluates a categorical loss across 16 classes. The network was trained with Adam optimizer, the learning rate was set to be $5\times10^{-4}$. In addition, similar to Lipsia, the team implemented a data augmentation strategy: samples from validation set have been added to the training data.

\textbf{Lipsia} \cite[Universität Leipzig]{lipsia2021sigtyp} submitted a constrained system based on the ResNet-50 \citep{he2016deep}, a  deep (50 layers) CNN-based neural architecture. The choice of the model is supported by a comparative analysis with more shallow architectures such as ResNet-34 and a 3-layer CNNs that all were shown to overfit to the training data. In addition, the authors proposed transforming MFCC features into corresponding 640x480 spectrograms since this data format is more suitable for CNNs. The output layer of the network is dense and evaluates  the probabilities of 16 language classes.%
\footnote{The submitted system actually predicts one out of 18 classes as two other languages that weren't part of the eventual test set were included. The system predicted these two languages for 27 of 8000 test examples, i.e., $\approx~0.34\%$.}
Finally, the authors augmented the training data with 60\% of the samples from the validation set because the training set did not present enough variety in terms of domains and speakers while the validation data included significantly more. Use of the validation data in this way seems to have greatly improved generalization ability of the model.  

The model performed relatively well with no  fine-tuning or transfer-learning applied after augmentation.\footnote{The authors trained ResNet-50 from scratch.}

\textbf{NTR} \cite[NTR Labs composite]{ntr2021sigtyp}, submitted an essentially constrained\footnote{Although technically external noise data was used when augmenting the dataset, no language-specific resources were.} system which uses a CNN with a self-attentive pooling layer. 
The architecture of the network was QuartzNet ASR following \citet{kriman2020quartznet}, with the decoder mechanism replaced with a linear classification mechanism. 
The authors also used a similar approach in another challenge on low-resource ASR, Dialog-2021 ASR\footnote{\url{http://www.dialog-21.ru/en/evaluation/}}. 
They applied several augmentation techniques, namely shifting samples in range (-5ms; +5ms), 
MFCC perturbations \cite[SpecAugment;][]{park2019specaugment}, 
and adding background noise.

\begin{table*}[tbh]
\newcommand{\famhead}{\cellcolor{gray!10}\color{black!50}}
\centering
\begin{adjustbox}{width=\linewidth}
\begin{tabular}{ccccccccccccc}
\toprule
\textbf{ISO} & \multicolumn{2}{c}{\textbf{Anlirika}} && \multicolumn{2}{c}{\textbf{Baseline}} && \multicolumn{2}{c}{\textbf{Lipsia}} && \multicolumn{2}{c}{\textbf{NTR}} \\
\cmidrule{2-3}\cmidrule{5-6}\cmidrule{8-9}\cmidrule{11-12}
 &  Valid. & Test &&  Valid. & Test && Valid. & Test && Valid. & Test \\
\midrule
\famhead\textit{Family: Afro-Asiatic} & \famhead\textit{.329} & \famhead.214 &\famhead& \famhead\textit{.181} & \famhead.235 &\famhead& \famhead\textit{.670} & \famhead\textbf{.453} &\famhead& \famhead\textit{.102} & \famhead.082 \\
kab        & \textit{.329} & .214 && \textit{.181} &         .235  && \textit{.670} & \textbf{.453} && \textit{.102} & .082 \\
\midrule
\famhead\textit{Family: Austronesian} & \famhead\textit{.429} & \famhead.368 &\famhead& \famhead\textit{.082} & \famhead.094 &\famhead& \famhead\textit{.578} & \famhead\textbf{.498} &\famhead& \famhead\textit{.065} & \famhead.060 \\
iba        & \textit{.692} & .696 && \textit{.029} &         .018  && \textit{.980} & \textbf{.968} && \textit{.020} & .031 \\
ind        & \textit{.350} & .108 && \textit{.033} &         .105  && \textit{.700} & \textbf{.338} && \textit{.096} & .074 \\
sun        & \textit{.406} & \textbf{.369} && \textit{.160} & .149 && \textit{.090} &         .140 && \textit{.086} & .082 \\
jav        & \textit{.267} & .300 && \textit{.106} &         .106  && \textit{.540} & \textbf{.547} && \textit{.059} & .053 \\
\midrule
\famhead\textit{Family: Basque} & \famhead\textit{.565} & \famhead.405 &\famhead& \famhead\textit{.100} & \famhead.090 &\famhead& \famhead\textit{.850} & \famhead\textbf{.792} &\famhead& \famhead\textit{.077} & \famhead.016 \\
eus        & \textit{.565} & .405 && \textit{.100} &         .090  && \textit{.850} & \textbf{.792} && \textit{.077} & .016 \\
\midrule
\famhead\textit{Family: Dravidian} & \famhead\textit{.351} & \famhead.246 &\famhead& \famhead\textit{.202} & \famhead.138 &\famhead& \famhead\textit{.807} & \famhead\textbf{.572} &\famhead& \famhead\textit{.074} & \famhead.053 \\
tam        & \textit{.342} & .272 && \textit{.348} &         .204  && \textit{.800} & \textbf{.609} && \textit{.172} & .046 \\
kan        & \textit{.188} & .168 && \textit{.000} &         .042  && \textit{.820} & \textbf{.557} && \textit{.004} & .015 \\
tel        & \textit{.523} & .298 && \textit{.259} &         .168  && \textit{.800} & \textbf{.550} && \textit{.046} & .097 \\
\midrule
\famhead\textit{Family: Indo-European} & \famhead\textit{.439} & \famhead.225 &\famhead& \famhead\textit{.130} & \famhead.144 &\famhead& \famhead\textit{.722} & \famhead\textbf{.402} &\famhead& \famhead\textit{.114} & \famhead.047 \\
hin        & \textit{.458} & .378 && \textit{.091} &         .099  && \textit{.780} & \textbf{.635} && \textit{.021} & .011          \\
por        & \textit{.211} & .143 && \textit{.157} &         .166  && \textit{.550} & \textbf{.358} && \textit{.102} & .068          \\
rus        & \textit{.630} & \textbf{.034} && \textit{.014} &         .014  && \textit{.900} & \textbf{.065} && \textit{.050} & \textbf{.049} \\
eng        & \textit{.194} & .148 && \textit{.161} &         .179  && \textit{.460} & \textbf{.414} && \textit{.270} & .099          \\
mar        & \textit{.701} & .423 && \textit{.229} &         .263  && \textit{.920} & \textbf{.539} && \textit{.126} & .010          \\
\midrule
\famhead\textit{Family: Niger-Congo} & \famhead\textit{.516} & \famhead.403 &\famhead& \famhead\textit{.138} & \famhead.063 &\famhead& \famhead\textit{.860} & \famhead\textbf{.763} &\famhead& \famhead\textit{.122} & \famhead.038 \\
cnh        & \textit{.516} & .403 && \textit{.138} &         .063  && \textit{.860} & \textbf{.763} && \textit{.122} & .038 \\
\midrule
\famhead\textit{Family: Tai-Kadai} & \famhead\textit{.362} & \famhead.156 &\famhead& \famhead\textit{.086} & \famhead.052 &\famhead& \famhead\textit{.780} & \famhead\textbf{.401} &\famhead& \famhead\textit{.025} & \famhead.015 \\
tha        & \textit{.362} & .156 && \textit{.086} &         .052  && \textit{.780} & \textbf{.401} && \textit{.025} & .015 \\
\midrule\midrule
F1, Macro Avg. & \textit{.421} & .282 && \textit{.131} &         .122  && \textit{.719} & \textbf{.508} && \textit{.086} & .049 \\
F1, Micro Avg. & \textit{.436} & .298 && \textit{.145} &         .137  && \textit{    } & \textbf{.532} && \textit{    } & .063 \\
\midrule
Accuracy   & & 29.9\% &&              &        13.7\% &&               & \textbf{53.1\%} &&             & 6.3\% \\
\bottomrule
\end{tabular}
\end{adjustbox}
\caption{F$_1$ scores, their macro-averages per family, and overall accuracies of submitted predictions on validation and test data (validation results are self-reported by participants). The Lipsia system performed best across nearly all languages and consistently achieves the highest averages.}
\label{tab:results-subm}
\end{table*}

\begin{figure*}[tbh]
    \centering
    \hspace*{-1em}
    \includegraphics[width=.95\linewidth]{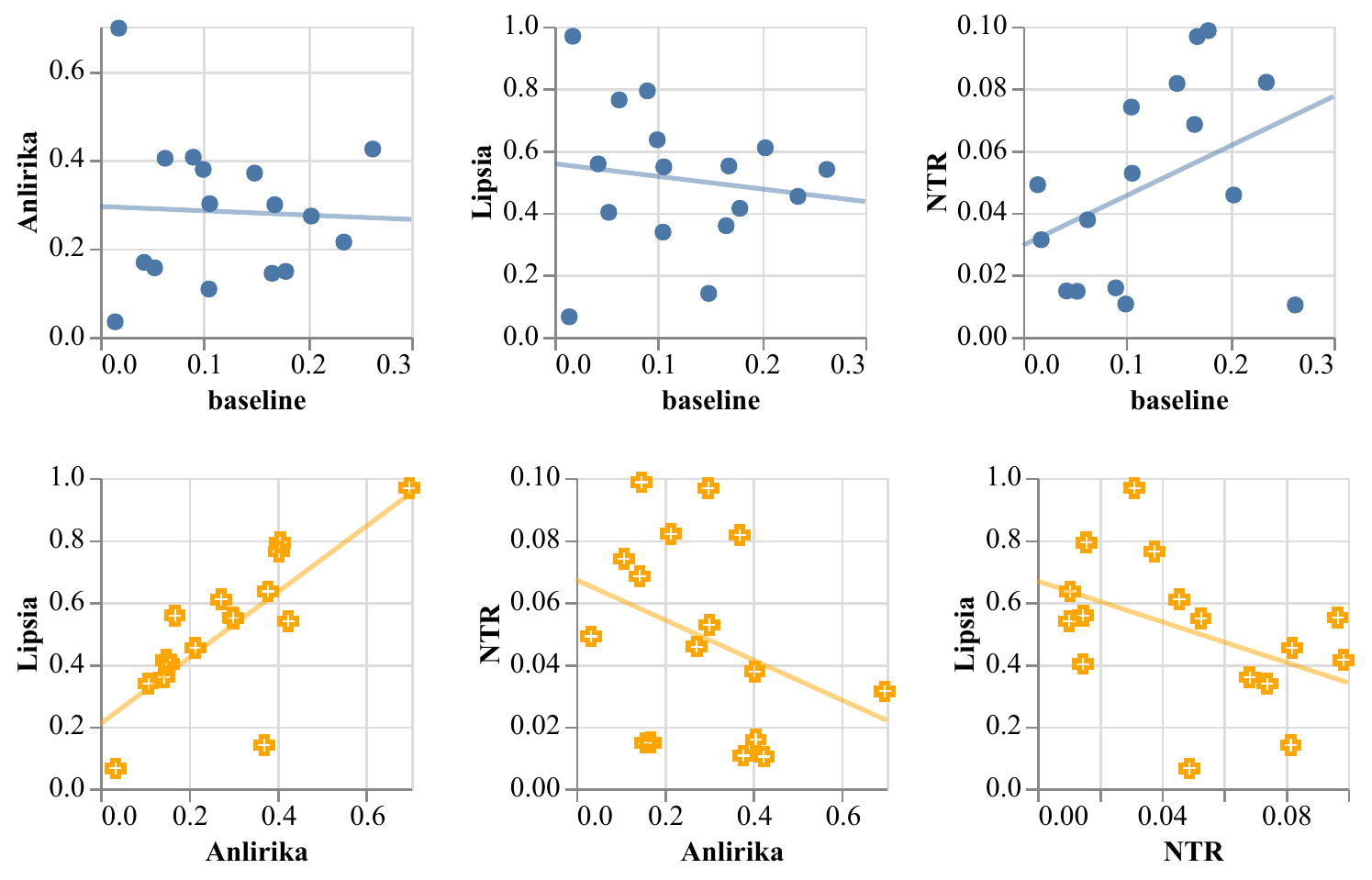}
    \caption{Correlating submitted systems' F$_1$ scores for our 16 languages on the test set. The lines are linear regressions as described in \autoref{sec:results}.}
    \label{fig:scatter-systems-against-baseline}
\end{figure*}

\section{Results and Analysis}\label{sec:results}

The main results in \autoref{tab:results-subm} show all systems greatly varying in performance, with the Lipsia system clearly coming out on top, boasting best accuracy and average F$_1$ score, and reaching the best F$_1$ score for nearly each language individually.%
\footnote{
Each of the ``wins'' indicated by boldface in \autoref{tab:results-subm} is statistically significant under a paired-permutation significance test (note that as we are not in a multiple-hypothesis testing setting, we do not apply Bonferroni or similar corrections). There are no significant differences between the baseline and the Anlirika system for kab, ind, por, rus, and eng; between the baseline and the Lipsia system for sun; between the baseline and the NTR system for ind, iba, and cnh; between Anlirika and Lipsia on rus; between Lipsia and NTR on rus; between Anlirika and NTR on ind and rus.
}

All four systems' performance varies greatly on average, but nevertheless some interesting overall trends emerge.
\autoref{fig:scatter-systems-against-baseline} shows that while the Anlirika and Lipsia systems' performance on the different languages do not correlate with the baseline system (linear fit with Pearson's $R^2=0.00$ and $p>0.8$ and $R^2=0.02$ and $p>0.5$, respectively), the NTR system's struggle correlates at least somewhat with the same languages that the baseline system struggles with: a linear fit has $R^2=0.15$ with $p>0.1$.
More interestingly, in correlations amongst themselves, the Anlirika and Lipsia systems do clearly correlate ($R^2=0.57$ and $p<0.001$), and the NTR system correlates again at least somewhat with the Anlirika system ($R^2=0.11$ and $p>0.2$) and the Lipsia system ($R^2=0.19$ and $p>0.05$).

Note that most systems submitted are powerful enough to fit the training data: our baseline achieves a macro-averaged F$_1$ score of $.98$ ($\pm .01$) on the training data, the Lipsia system similarly achieves $.97$ ($\pm .03$), the NTR system reaches a score of $.99$ ($\pm .02$). An outlier, the Anlirika system reaches only $.75$ ($\pm .09$).
On held-out data from CMU Wilderness which matches the training data domain, the baseline achieves $.96$ F1. 
This suggests an inability to generalize across domains and/or speakers without additional data for adaptation. 

\begin{figure*}
    \centering
    \includegraphics[width=\linewidth]{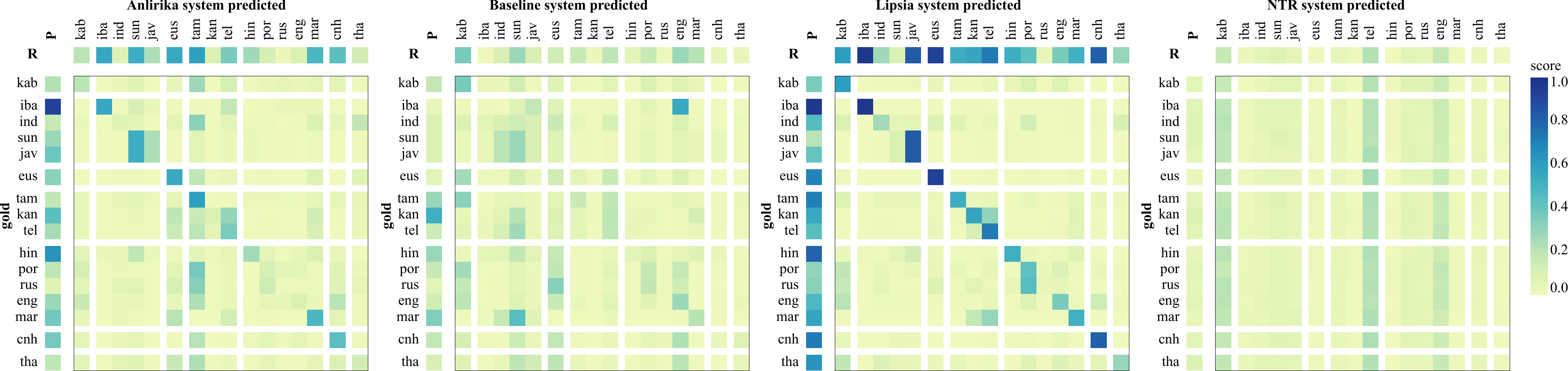}
    \caption{Visualization of Precision (P), Recall (R), and confusion matrices (scores are counts normalized by number of gold entries) for the Anlirika, baseline, Lipsia, and NTR system, grouped by language families.}
    \label{fig:confusion}
\end{figure*}

Diving deeper into performance on different languages and families, \autoref{fig:confusion} shows confusion matrices for precision and recall, grouped by language family. We can see the superiority of the Lipsia system and to a lesser degree the Anlirika system over the generally more noisy and unreliable baseline system and the NTR system which was likely overtrained: it classifies 23\% of examples as tel, 20\% as kab, and 16\% as eng, with the remaining 41\% spread across the remaining 13 languages (so $\approx$ 3.2\% per language).

Interestingly, the other three systems all struggle to tell apart sun and jav, the Anlirika and baseline systems classifying both mostly as sun and the Lipsia system classifying both mostly as jav. Note that the baseline system tends to label many languages' examples as sun (most notably mar, the test data for which contains only female speakers), eus (most notably also rus), and eng (most notably also iba), despite balanced training data. In a similar pattern, the Anlirika predicts tam for many languages, in particular ind, the other two Dravidian languages kan and tel, por, rus, eng, cnh, and tha.

Looking more closely at the clearly best-performing system, the Lipsia system, and its performance and confusions, we furthermore find that the biggest divergence from the diagonal after the sun/jav confusion is a tendency to label rus as por, and the second biggest divergence is that mar examples are also sometimes labeled as kan and tel; while the first one is within the same family, in the second case, these are neighbouring languages in contact and mar shares some typological properties with kan (and kan and tel belong to the same language family).

\section{Conclusion}

This paper describes the SIGTYP shared task on robust spoken language identification (SLID). 
This task investigated the ability of current SLID models to generalize across speakers and domains. 
The best system achieved a macro-averaged accuracy of 53\% by training on validation data, indicating that even then the task is far from solved.
Further exploration of few-shot domain and speaker adaptation is necessary for SLID systems to be applied outside typical well-matched data scenarios.

\bibliography{bibliography,custom}
\bibliographystyle{acl_natbib}

\end{document}